\documentclass[letterpaper, 10 pt, conference]{ieeeconf}  % Comment this line out
\IEEEoverridecommandlockouts                              % This command is only
\usepackage{url}
\usepackage{times}
\usepackage{helvet}
\usepackage{courier}
\usepackage{amssymb}
\usepackage{amsmath}
\usepackage{multirow}
\usepackage{array}
\usepackage{hhline}
\usepackage{graphicx}
\usepackage[super]{nth}
\usepackage{bm}

\usepackage[colorinlistoftodos]{todonotes}
\usepackage{subfig}

\title{\LARGE \bf
Generative Adversarial Network Rooms in Generative Graph Grammar Dungeons for \emph{The Legend of Zelda}
}

\author{Jake Gutierrez$^{1}$ and Jacob Schrum$^{2}$% <-this % stops a space
\thanks{$^{1}$J. Gutierrez is an undergraduate at Southwestern University,
        Georgetown, TX 78626, USA
        {\tt\small gutierr8@southwestern.edu}}%
\thanks{$^{2}$J. Schrum is an Assistant Professor of Computer Science at Southwestern University,
        Georgetown, TX 78626, USA
        {\tt\small schrum2@southwestern.edu}}%
}

\begin{document}

\maketitle
\thispagestyle{empty}
\pagestyle{empty}

%%%%%%%%%%%%%%%%%%%%%%%%%%%%%%%%%%%%%%%%%%%%%%%%%%%%%%%%%%%%%%%%%%%%%%%%%%%%%%%%
\begin{abstract}

Generative Adversarial Networks (GANs) have demonstrated their ability to learn patterns in data and produce new exemplars similar to, but different from, their training set in several domains, including video games. However, GANs have a fixed output size, so creating levels of arbitrary size for a dungeon crawling game is difficult. GANs also have trouble encoding semantic requirements that make levels interesting and playable. This paper combines a GAN approach to generating individual rooms with a graph grammar approach to combining rooms into a dungeon. The GAN captures design principles of individual rooms, but the graph grammar organizes rooms into a global layout with a sequence of obstacles determined by a designer. Room data from \emph{The Legend of Zelda} is used to train the GAN. This approach is validated by a user study, showing that GAN dungeons are as enjoyable to play as a level from the original game, and levels generated with a graph grammar alone. However, GAN dungeons have rooms considered more complex, and plain graph grammar's dungeons are considered least complex and challenging. Only the GAN approach creates an extensive supply of both layouts and rooms, where rooms span across the spectrum of those seen in the training set to new creations merging design principles from multiple rooms.

\end{abstract}

%%%%%%%%%%%%%%%%%%%%%%%%%%%%%%%%%%%%%%%%%%%%%%%%%%%%%%%%%%%%%%%%%%%%%%%%%%%%%%%%
\section{Introduction}

%% Don't think we need footnote for Rogue after all
% \footnote{\url{https://yukkurigames.com/pphs/}}

Video game developers increase replayability and reduce costs using Procedural Content Generation (PCG \cite{noor2016pcg}). Instead of experiencing the game once, players see new variations on every playthrough. %Procedurally generated levels allow for significant and obvious differences. 
This concept was introduced in \emph{Rogue} (1980), which procedurally generates new dungeons on every play. PCG is also applied to modern games like \emph{Minecraft} (2009), where users play on generated landscapes, and \emph{No Man's Sky} (2016), where procedurally generated worlds contain procedurally generated animals. PCG encourages increased exploration and increases replayability. 

An emerging PCG technique is Generative Adversarial Networks (GANs \cite{goodfellow:nips14}) used to search the latent design space of video game levels, as has been done in \emph{Super Mario Bros.} \cite{volz:gecco2018}, \emph{Doom} \cite{giacomello:cog19}, an educational game \cite{park:cog19}, and the General Video Game AI (GVG-AI \cite{PerezLiebana:tog19gvgai}) adaptation of \emph{The Legend of Zelda} \cite{Torrado2019BootstrappingCG}. In the GVG-AI version of Zelda, single-room levels require the player to fight enemies, reach a key, and take it to the exit. The technique applied by Torrado et al.~\cite{Torrado2019BootstrappingCG} to this game focuses on modeling non-local dependencies with the GAN in order to assure functional placement of the key and the exit door.
Their work addresses a problem GANs have with learning level semantics, but the levels are restricted in scale based on the size of training instances.

This paper explores a new hybrid PCG approach for dungeon crawlers based on levels from the actual \emph{Legend of Zelda} (1986). Specifically, a GAN generates rooms based on the Video Game Level Corpus (VGLC \cite{summerville:vglc2016}) description of the game. To scale up to large dungeons with interesting challenges, rooms are organized into a dungeon using a generative graph grammar \cite{dormans:pcg10} which maps a high-level, human-designed \emph{mission} to a sequence of room obstacles, and ultimately a complete dungeon. Combining the techniques creates new and interesting dungeons of arbitrary size.

This new technique (\texttt{Graph+GAN}) was evaluated by 30 human subjects. Each played three types of dungeons to compare the enjoyability, complexity, novelty, organization, and challenges of each through surveys. They played a dungeon from the original \emph{Legend of Zelda}, a graph grammar dungeon with rooms from the original game, and a dungeon generated with the new \texttt{Graph+GAN} technique.
Players rated dungeons roughly the same in most metrics. The exception is that GAN rooms were significantly less organized, and were considered most complex by a significant number of participants. 

%% Not a significant result ... this detail can wait
%Although most participants actually preferred the level from the original game most, the number was not significant.

These findings show that this technique can generate levels similar to hand-crafted dungeons from \emph{The Legend of Zelda}. However, these dungeons also contain unique new content, and a multitude of such dungeons can be generated.

%%% As usual, no space for the roadmap

%This paper covers Related Work (Section \ref{section:related}) followed by details on \emph{The Legend of Zelda} (Section \ref{section:loz}). Discussion about the GAN, graph grammar, and combining the two to make a playable game is covered in Section \ref{section:dungeon}. Next comes details on the Human Subject Study (Section \ref{section:hss}), its results (Section \ref{section:results}), and Discussion and Future Work (Section \ref{section:discussionAndFW}) before concluding (Section \ref{section:conclusion}).

%%%%%%%%%%%%%%%%%%%%%%%%%%%%%%%%%%%%%%%%%%%%%%%%%%%%%%%%%%%%%%%%%%%%%%%%%%%%%%%%
\section{Related Work} \label{section:related}

Procedural dungeon generation has been a topic of interest since \emph{Rogue} was released in 1980. As more complex games were released, the idea of procedurally generating dungeons became more prevalent. The popular games in the \emph{Diablo} series use PCG for generating dungeons, quests, and events. These features add variety and make these games more interesting and unpredictable, increasing replayability.

Procedural generation of dungeons has been widely studied in academia \cite{linden:tciaig13survey}.
Some representative techniques include cellular automata \cite{johnson:pcg10automata}, various evolutionary approaches \cite{valtchanov:c3s2e12,ashlock:tciaig11}, and generative grammars \cite{dormans:pcg10,dormans:tciaig11}. 

Dormans used a generative graph grammar to procedurally generate a dungeon mission, and a shape grammar to generate the dungeon itself \cite{dormans:pcg10}. Graph and shape grammars were further explored to generate dungeons similar to \emph{The Legend of Zelda: A Link to the Past} (LttP) in an undergraduate thesis \cite{Lavender2016TheZD}. These dungeons required particular graph and shape grammars to produce results similar to LttP. Although new dungeon layouts were created, the rooms came from LttP rather than being generated from scratch. 

A recent development is the use of GANs to model the latent design space of a level corpus. Volz et al.\ used a GAN to generate \emph{Super Mario} levels with objective-based evolution \cite{volz:gecco2018}. A similar approach was later applied to \emph{Doom} levels \cite{giacomello:cog19}. A GAN can even be replaced with an autoencoder, as was done to evolve levels for \emph{Lode Runner} \cite{thakkar:cog2019}. The approach worked in Mario despite a small data set, and the \emph{Doom} and \emph{Lode Runner} data sets were quite large. 

However, for certain games it is hard to produce playable levels because of limited training data. This challenge was overcome by Park et al.\ with multiple GANs \cite{park:cog19}: one GAN to create levels for a puzzle game from a small training set, and a second GAN using an augmented data set consisting of the original set plus levels from the first GAN that were actually solvable. Torrado et al.\ \cite{Torrado2019BootstrappingCG} used a similar approach, incorporating playable levels back into the training set when designing levels for the GVG-AI \cite{PerezLiebana:tog19gvgai} version of Zelda.

%There have been other research using GANs to generate levels such as generating levels from Doom that also incorporates a CMA-ES \cite{giacomello:cog19}. An educational puzzle game that generated levels from a GAN that is able to adapt to the player's skill \cite{park:cog19}. Thakkar et al. worked on a way to use an Autoencoder paired with a GAN to generate levels from Lode Runner \cite{thakkar:cog2019}.

%By combining Procedural Content Generation and AI, we are able to make dungeon layouts using Graph Grammar and then use a GAN to generate rooms within the dungeon.

In this paper, rather than make the GAN do more work, a division of labor is imposed. The GAN models the interior of individual rooms, and a generative graph grammar determines the dungeon layout and what items/obstacles are placed in each room. The result is a method that creates dungeons based off of \emph{The Legend of Zelda}, described next. 

\begin{figure}[t!]
\centering
\includegraphics[width=\columnwidth]{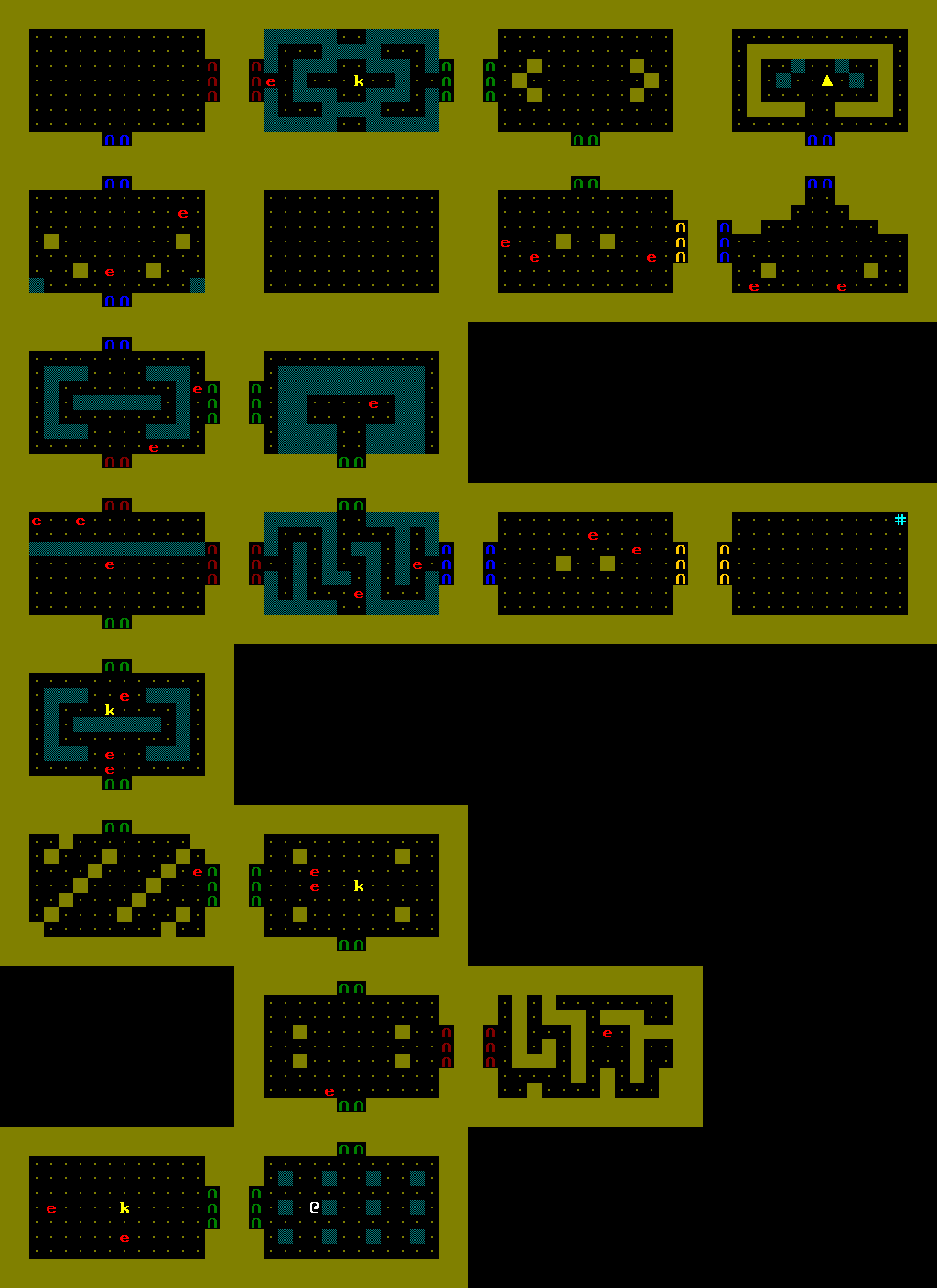}
\caption{Dungeon 4-1 from \emph{Legend of Zelda} converted to the Rogue-like engine. The goal is to reach the \emph{Triforce} (triangle) in the top-right room. In the middle right, there is a blue \# item, which is the raft; it allows players to cross one water tile (dark blue), which is necessary to traverse the room three spaces to the left of the room with the raft. In the original game, the raft room was underground (via stairs), and did not appear on the map. Due to limitations of the game engine for this study, the room was directly added to the map.}
\label{fig:tloz_4_1}
\end{figure}

\section{The Legend of Zelda} \label{section:loz}

%\todo[inline]{Add content here and also move some content from other sections here}

\emph{The Legend of Zelda} involves 18 dungeons across two quests (9 each) accessible via an overworld map. Each dungeon is composed of several rooms filled with enemies, items, and secret passages, where the end goal of each dungeon is find a \emph{Triforce}, which completes the dungeon. 

Each room is the same size. Although room layouts vary, many are reused both within and across dungeons. Rooms can be connected in a variety of ways: simple doors, doors requiring a key (\emph{Lock}), doors that only open when all enemies in the room are defeated (\emph{Soft Lock}), doors that open when a puzzle is solved (\emph{Puzzle}), and passages that need to be bombed to open. These connections are always in a side wall of the room, though some dungeons have stairs to standalone rooms that are not part of the main map layout. Stairs are excluded from dungeons in this study.

Many interesting items can be collected in the game, but only a few are relevant to this paper: keys, hearts, bombs, and the raft. 
Hearts replenish a player's health. Bombs allow the player to blow up walls to reveal hidden doors or kill enemies. The raft item allows players to move across one water tile. It is introduced in Dungeon 4-1 (4th dungeon of Quest 1, Fig.~\ref{fig:tloz_4_1}) and used throughout the rest of the game. 

%%% I just added parenthetical mentions of these labels to the paragraph above
%\emph{Soft lock} rooms also have enemies, but trap the player in the room until all enemies are defeated. \emph{Puzzle} rooms are also present and require the player to push a block to open a stairway.

Data about Zelda levels was obtained from the Video Game Level Corpus (VGLC \cite{summerville:vglc2016}). This data provides text representations of the tiles present in each dungeon. Details of this representation, and how it maps to the one used in this paper, are in Table~\ref{tab:tiles}. There are many symbols from the VGLC data, but since many of these tiles serve the same purpose as others, the tile training set is simplified.

%%% I had forgotten that some statues attack, but since that is rare (and not present in D4-1 I believe) let's just not mention it.
%with a few exceptions such as the Monster Statue and Stairs. In some dungeons, the Statue can shoot fire balls directly at the player, but this is not present in the rogue-like and is replaced with the water tile. Stairs can lead the player to an underground section of the dungeon, but is left out for simplicity of dungeon generation.

%%%%%%%%%%%%%%%%%%%%%%%%%%%%%%%%%%%%%%%%%%%%%%%%%%%%%%%%%%%%%%%%%%%%%%%%%%%%%%%%
\section{Dungeon Generation} \label{section:dungeon}

A GAN is trained to generate individual rooms, which can then be combined into dungeons using a generative graph grammar. The 2D layout of the rooms is derived in part from the graph. To assure that the dungeon is beatable, some additional walls may need to be knocked down. Users can then play a Rogue-like game in the repaired dungeon.

\begin{table}[t!]
\caption{\label{tab:tiles}Tile Types Used in Generated Zelda Rooms.}
{\small Tile types come from VGLC, 
but many were unnecessary in the simplified Rogue-like engine used to play the levels. Thus the available tile set was reduced to three relevant types: floor, wall, and water. VGLC rooms were converted to use only these three tile types when serving as training input to the discriminator, and GAN outputs were used to make rooms using only these three tiles.}

\centering
\begin{tabular}{c|c|c|c|c}
\hline
Tile type & VGLC & Game & Rogue-like & Rogue Type\\
\hline % Images needed to be vertically centered within row
Floor & F & \includegraphics[width=10pt]{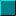} & \includegraphics[width=10pt]{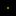} & Floor \\
Wall & W & \includegraphics[width=10pt]{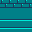} &  \includegraphics[width=10pt]{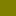} & Wall \\
Block & B & \includegraphics[width=10pt]{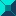} & 
\includegraphics[width=10pt]{rouge-wall.png}& Wall \\
Door & D & \includegraphics[width=10pt]{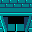} &  \includegraphics[width=10pt]{rouge-wall.png}& Wall \\
Stair & S & \includegraphics[width=10pt]{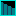} &  \includegraphics[width=10pt]{rouge-wall.png}& Wall \\
Water & P & \includegraphics[width=10pt]{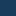} &  
\includegraphics[width=10pt]{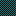}& Water \\
Walk-able Water & O & \includegraphics[width=10pt]{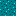} &  \includegraphics[width=10pt]{rouge-water.png}& Water\\
Water Block & I & \includegraphics[width=10pt]{block.png} &  \includegraphics[width=10pt]{rouge-water.png}& Water\\
Monster Statue & M & \includegraphics[width=10pt]{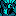} &  \includegraphics[width=10pt]{rouge-water.png}& Water\\

\hline
\end{tabular}
\end{table}

\subsection{Zelda GAN} \label{section:gan}

\begin{figure}[t]
\centering
\includegraphics[width=\columnwidth]{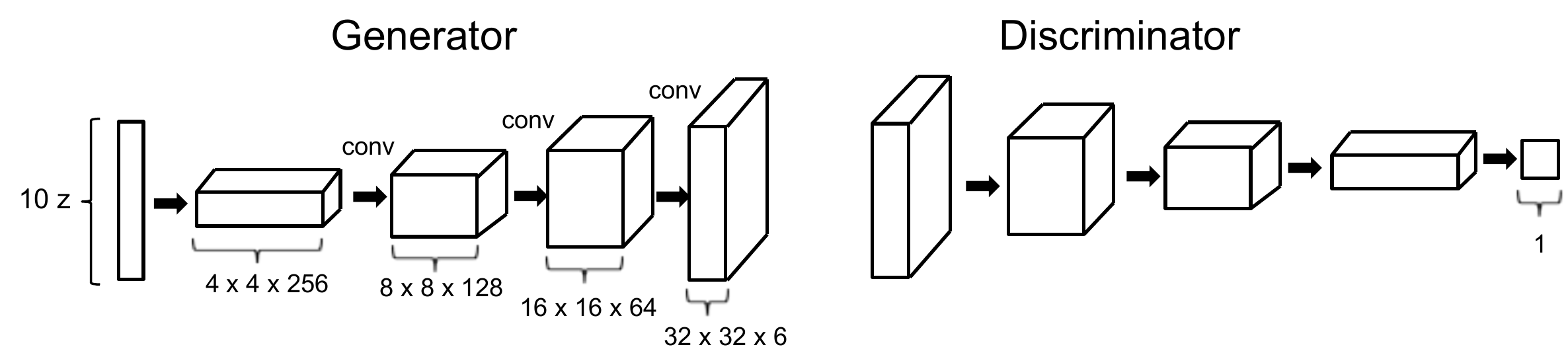}
\caption{The Zelda GAN architecture.}
\label{fig:architecture}
\end{figure}

To generate Zelda rooms, the same GAN architecture/code used in Mario \cite{volz:gecco2018} is used (Fig.~\ref{fig:architecture}). The only differences are a change in output size to accommodate a different tile type count, and a reduced latent vector size of 10 because initial experiments indicated that an unnecessarily large latent vector led to large areas in the latent space with little variation. The output width and height were maintained at %the unnecessarily large size of 
$32 \times 32$ for backwards compatibility. Zelda rooms are only $16 \times 11$, but the GAN makes the surrounding space floor tiles.

This GAN can be trained on any 2D tile-based level representation.
The generator takes latent vectors of noise from $[-1,1]^{10}$ as input, and outputs a 3D volume of $32 \times 32$ vectors of length $6$. Each value in each vector corresponds to a tile type in Table \ref{tab:tiles}, and these vectors are collapsed so that the tile at its position in the resulting 2D image corresponds to the maximum value in the vector\footnote{Only three of six values are used. The GAN originally supported six tile types, but this setting was not changed after settling on three tile types.}. %Essentially, an \texttt{argmax} operation is performed across all of the vectors. 
The upper-left $16 \times 11$ portion of the image can then be interpreted as a Zelda room.

An additional discriminator network is also used during training. Its input is a one-hot encoded version of either a Zelda room from the training set, or fake output produced by the generator. Over the course of 10,000 epochs it is trained to make its single output $1$ for real Zelda rooms and $-1$ for generated rooms. The generator itself is trained along with the discriminator, to the point where it produces convincing fake Zelda rooms. After training, the discriminator performs no better than a coin toss, and is thus discarded.

%. The discriminator compares the original training data with data generated by the generator. The GAN takes the discriminator's information and improves itself with each iteration. This process reoccurs until a set number of iterations, 10000 (epochs = 10000) was used for this model. The discriminator is then discarded and only the generator is used to retrieve generated rooms. To explore the generator, a latent vector of size 10 was used. To generate GAN room, the trained GAN model is given an array of ten randomly choosen doubles, the latent vector. 

To generate the training set, the 18 dungeons in VGLC were split into rooms and encoded as GAN inputs. Because there are many repeated rooms throughout the dungeons, duplicates were eliminated. Some Zelda tiles have a similar function, but a different aesthetic. Such tiles were merged into one, as seen in Table \ref{tab:tiles}. The VGLC data incorrectly designates statue tiles as monsters, but the GAN interprets them as water tiles. %, though block tiles would also have been appropriate. 
Additionally, doors were removed from the training data, because doors need to be placed in accordance with the game mission defined by the graph grammar.

\subsection{Graph Grammar} \label{section:graphgrammar}

\begin{figure}[t]
\centering
\includegraphics[width=\columnwidth]{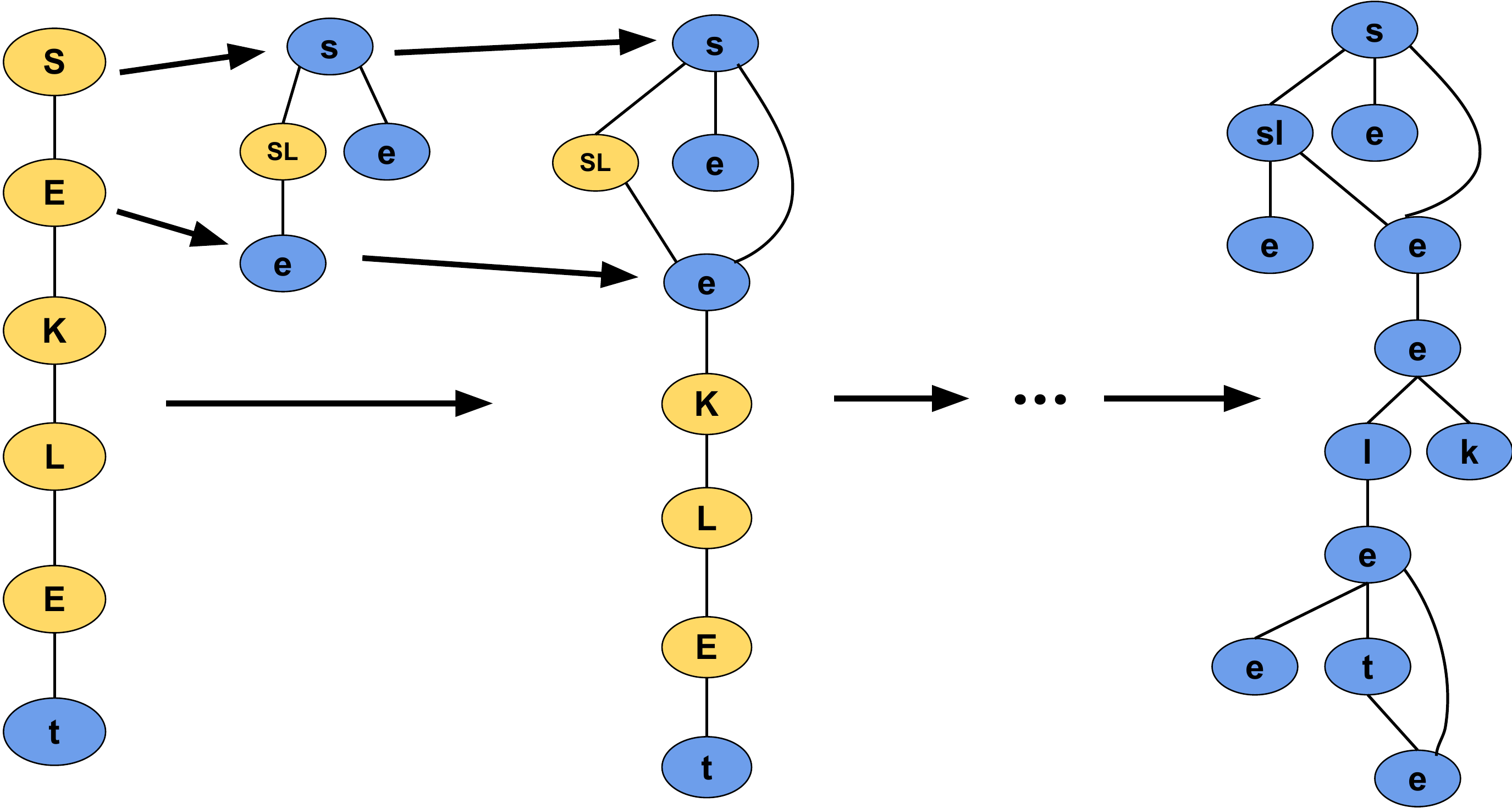}
\caption{Graph Expansion Example. The first two nodes of the initial graph are replaced with a randomly chosen sub-graph defined by the available graph grammar rules. Non-terminal symbols are represented as capital letters (yellow) and terminals as lower case letters (blue). The process repeats until there are no non-terminals. Nodes correspond to rooms.}
\label{fig:graph_grammar}
\end{figure}

\begin{figure*}[p]
    \centering
    
\includegraphics[width=\textwidth]{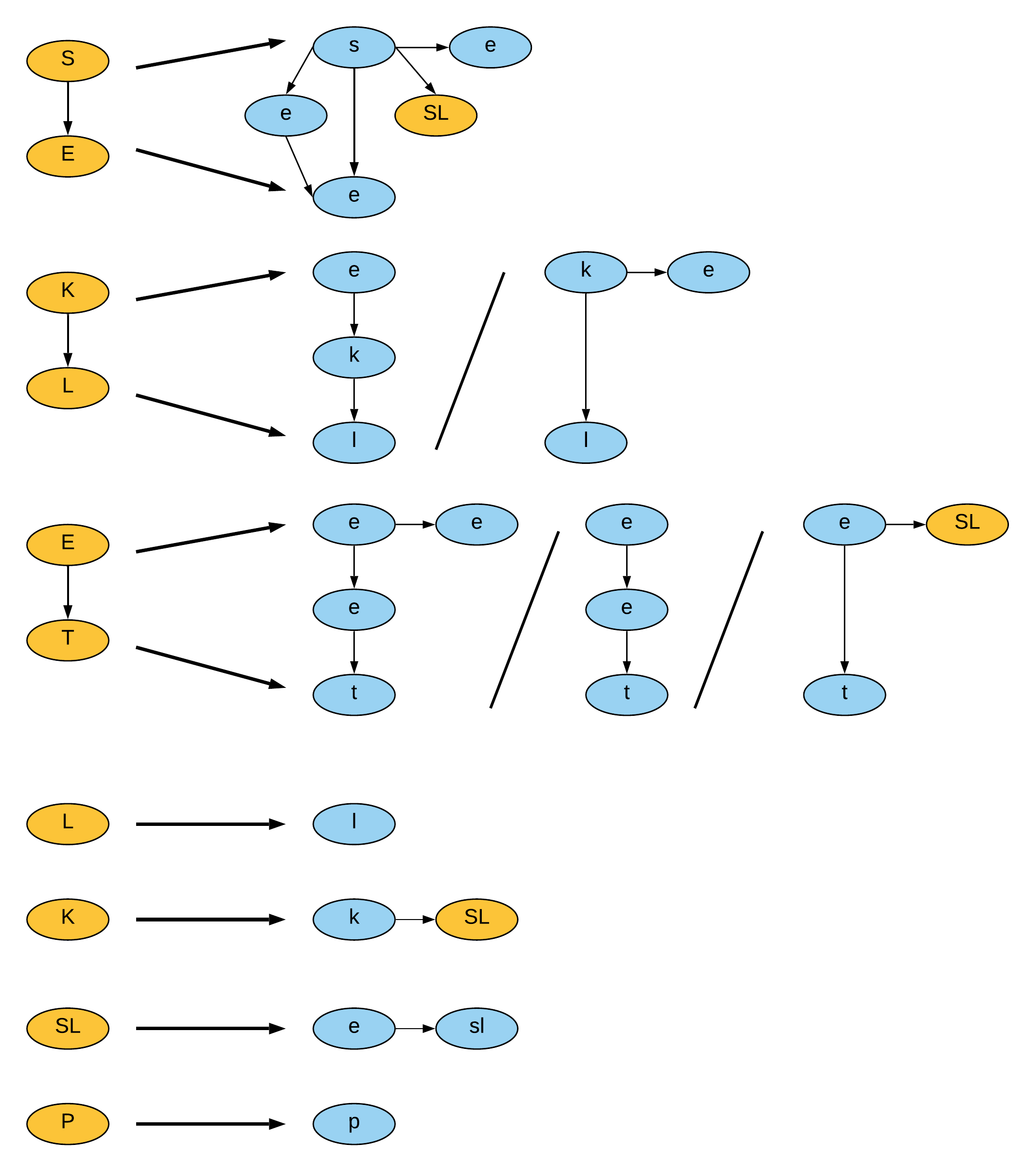}
\caption{Graph Grammar Rules. This set of rules defines how symbols (Orange/Uppercase) can map to terminals (Blue/Lowercase) in a final dungeon. Though the set is small, and some symbol pairs only have one possible transformation, there is enough variety in the rule set to create many different dungeons, especially when combined with different room placements and layouts.}
\label{fig:graphGrammarSupp}
\end{figure*}

A generative graph grammar \cite{dormans:pcg10} determines how rooms connect in a dungeon. 
A designer-provided \emph{backbone} graph represents the \emph{mission} of a dungeon. The backbone includes specific rooms that must be present in the dungeon. The backbone used in this paper is
\emph{Start} $\rightarrow$ \emph{Enemy} $\rightarrow$ \emph{Key} $\rightarrow$ \emph{Lock} $\rightarrow$ \emph{Enemy} $\rightarrow$ \emph{Key} $\rightarrow$ \emph{Puzzle} $\rightarrow$ \emph{Lock} $\rightarrow$ \emph{Enemy} $\rightarrow$ \emph{Triforce}. The backbone is a sequence of non-terminal symbols that get replaced until only terminals remain. %While this backbone is linear, the designer can implement any type of graph as the starting point. 
For each pair or single symbol there is a finite set of rules defining what could replace it.
For example, (\emph{Key} $\rightarrow$ \emph{Lock}) 
could be replaced with % [Where did this example come from? ->] a starting room that has two adjacent empty rooms and one adjacent enemy room, which leads to the rest of the dungeon. 
a key room that has two neighbors: a dead-end enemy room, and a locked room leading onward.
Full rule set is in online material\footnote{\url{southwestern.edu/~schrum2/zeldagan.html}} (Fig.\ \ref{fig:graphGrammarSupp}).
Each rule defines a mini-graph that is placed into the backbone and can be made up of both non-terminals and terminals. An example of the iterative replacement process is in Fig.~\ref{fig:graph_grammar}. This process can generate multiple graphs representing different dungeons, but ensures that the general sequence stays the same.

%% There are a total of 16 rules in the grammar, though more could easily be added.

\setlength\tabcolsep{2pt}
\begin{table}[t]
\caption{\label{tab:nonTerminals}Non-terminal Graph Grammar Symbols.}
{\small Each non-terminal symbol defines a type of room that must be in the dungeon, but during the generation process, edges connecting to non-terminal symbols get transformed to more elaborate sub-graphs that contain terminal representations of indicated rooms.} 

\centering
\begin{tabular}{|r|c|p{0.75\columnwidth}|}
\hline
Symbol & Short & Description \\
\hline 
\emph{Start} & \emph{S} & Dungeon starting room. Only one. \\
\hline 
\emph{Enemy} & \emph{E} & Room with random number of enemies. \\
\hline
% \emph{Nothing} & \emph{N} & No added content. \\
% \hline 
\emph{Key} & \emph{K} & Has enemies, and key appears after defeating them. \\
\hline 
\emph{Lock} & \emph{L} & Has door that is unlocked by a key. \\
\hline 
\emph{Soft Lock} & \emph{SL} & Has enemies, and a door that opens when they are defeated. Also contains raft item. \\
\hline 
\emph{Puzzle} & \emph{P} & Has door that opens when puzzle block is pushed. \\
\hline 
\emph{Triforce} & \emph{T} & Has Triforce. Dungeon complete once found. Only one. \\
\hline 
\end{tabular}
\end{table}

Non-terminals used by the grammar are in Table \ref{tab:nonTerminals}. Not all symbols in the table are in the initial backbone, but can be added by grammar rules. The available rules assure that at least one \emph{Soft Lock} room is in every dungeon, despite its absence from the backbone.
Once a graph is created, the actual 2D layout of rooms must be determined.

%%% The Table expalins these details now
%The graph grammar uses the following non-terminals. \emph{Start} is the room where the player starts in the dungeon. \emph{Enemy} rooms contain 1 to 3 enemies randomly spawned in a room. The soft lock room, as mentioned previously, was in the original Legend of Zelda, so we included this feature in the graph, and there is a guarantee to be one \emph{Soft lock} room in each graph generated. \emph{Lock} rooms have a locked door that requires a key in order to progress. There are two \emph{Lock} rooms in our graph grammar. A key can be obtained in a \emph{Key} room, where a key appears once all of the enemies have been defeated in that room. By design, every \emph{Lock} room has a \emph{Key} room right before it in comparison to where the \emph{Start} room is in the dungeon. Finally, there is the \emph{Triforce} room, where the Triforce is placed and is the final goal of the dungeon.

\subsection{Dungeon Layout} \label{section:dungeonlayout}
%TODO: Describe process of putting GAN and Graph together
Dungeon rooms are placed in breadth-first order beginning with the start room of the graph. However, there may not be space around a room to accommodate its neighbors. To ensure that all rooms are placed, the algorithm backtracks if no space is available around a room needing a neighbor. 

\begin{figure}[t]
    \centering
    
% \begin{subfigure}[.5\textwidth]
%     \centering
%     \includegraphics[width=\linewidth]{dungeon_11_no-test-graph-overlay.pdf}
%     \caption{Graph Representation of Dungeon}
%     \label{fig:generated_dungeon_graph}
% \end{subfigure}
% \begin{subfigure}[.5\textwidth]
%     \centering
%     \includegraphics[width=\linewidth]{dungeon_11_no-test.png}
%     \caption{Corresponding 2D Layout of Dungeon}
%     \label{fig:generated_dungeon_graph}
% \end{subfigure}
\makebox[\columnwidth]{
    \subfloat[Graph Representation of Dungeon]{
        \includegraphics[width=\columnwidth]{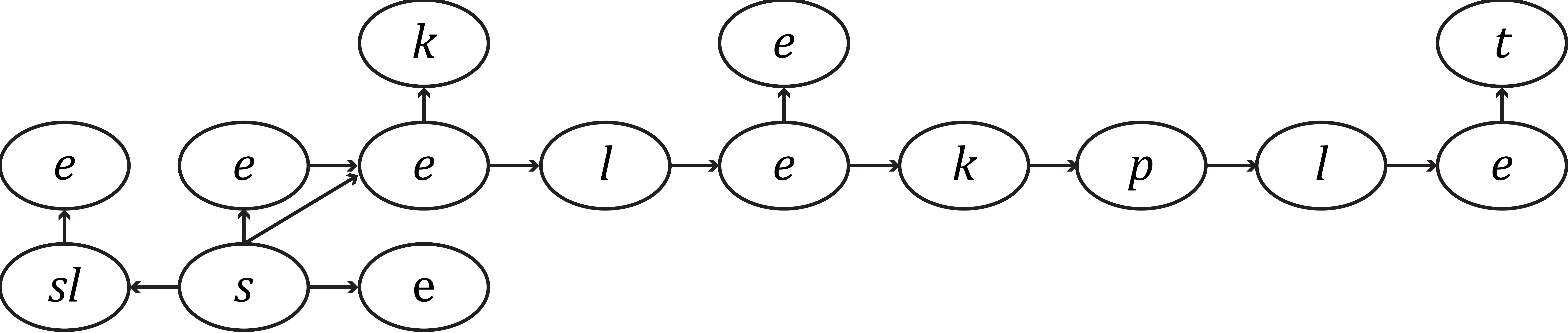}
        \label{fig:generated_dungeon_graph}}
}
\makebox[\columnwidth]{
    \subfloat[Corresponding 2D Layout of Dungeon]{
        \includegraphics[width=\columnwidth]{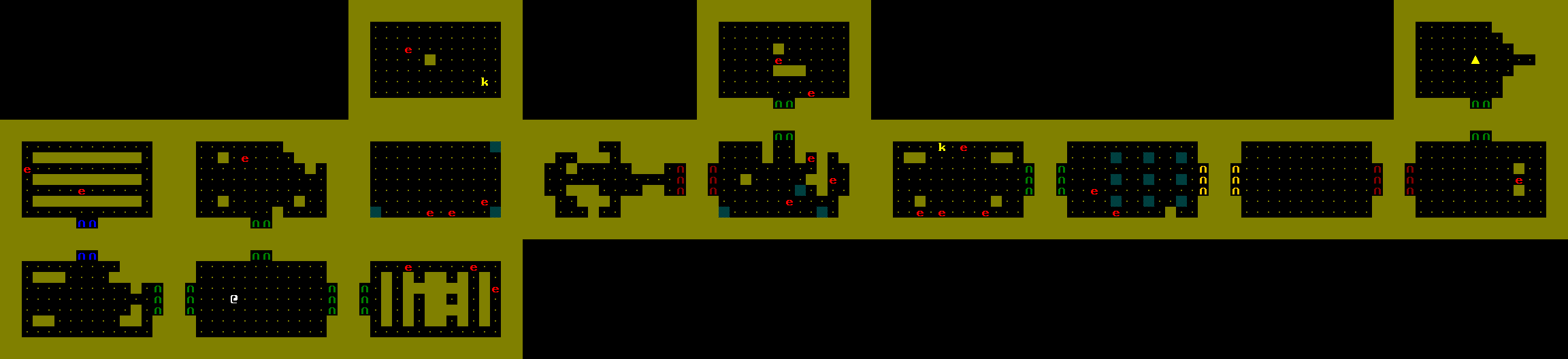}
        \label{fig:generated_dungeon}}
}
\caption{Creation of Dungeon From Graph. \protect\subref{fig:generated_dungeon_graph} Graph that represents a dungeon. Each node represents a room, and each edge represents a doorway between rooms. Symbols in each node indicate the type of obstacles present in the room. The graph is directed, but the player can go back and forth between rooms. The directed edges show how the player would encounter each room for the first time. \protect\subref{fig:generated_dungeon} The complete generated dungeon based on the above graph, with specific room layouts determined by the GAN. Note that one edge of the graph (diagonal from \emph{s} to \emph{e}) is lost when generating the dungeon.}
\label{fig:graphAnd2D}
\end{figure}

Specifically, a list of \emph{edges} (between rooms) is generated in breadth-first order from the start room. This list is iterated through, and any room in an edge not yet placed is added to the dungeon. After the start room, all rooms must be placed in relation to the first room in an edge.
A random position orthogonally adjacent to the previously placed room is chosen for its neighbor, backtracking whenever
all surrounding positions of a room are occupied. Backtracking undoes the last placement and attempts an alternative that has not yet been tried.
Search continues until the list of edges is exhausted. Note that only the first occurrence of each room is placed.
Although the graph represents the connectivity of rooms, the 2D layout typically loses edges present in the original graph.
The layout attempts to match the graph as closely as possible (Fig.~\ref{fig:graphAnd2D}).

Certain rooms are manipulated according to the grammar symbols. \emph{Enemy} rooms randomly get 1--3 enemies placed in random locations.
\emph{Key} rooms have a key placed in a random empty spot in the room followed by randomly placed enemies. \emph{Lock} rooms have a locked door placed at the connection leading to the next room. \emph{Soft Lock} rooms have a soft locked door and randomly placed enemies. Additionally, the first soft locked room of the dungeon contains a raft item. 
\emph{Puzzle} rooms have a door that can only be opened by finding and pushing a particular block in the room in a specific direction. A random spot in the room currently with or without a block becomes the puzzle block.
\emph{Triforce} room has a Triforce, represented as a yellow triangle, in the middle of the room. Bomb-able doors have a 40\% chance of replacing a normal door; normal meaning that it is not locked, soft locked, or puzzle locked. 

\subsection{Room Repair} \label{section:repair}

To assure that each dungeon is beatable, some rooms are modified to create a path between certain points of interest. A* search is used to check that dungeons are beatable. The A* state representation tracks puzzle blocks, keys, and the raft item, but ignores enemies and always assumes there are sufficient bombs for bomb-able walls.
%An A* agent was already created before to determine whether a dungeon was beatable or not. Using the data from an A* agent that was unable to beat our generated dungeon, we can 

If A* fails to beat a dungeon, then one room is modified. Each room has points of interest (POIs): doors, keys, puzzle blocks, the raft, and the Triforce. A* tracks the visited and unvisited POIs. On search failure, a random unvisited POI is chosen along with a random visited POI in the same room
(if there were no visited POIs, then two unvisited POIs are chosen). A modified Bresenham's line algorithm \cite{bresenham:ibm65} draws floor tiles from the visited POI to the unvisited POI. Puzzle blocks are a special case requiring POIs for both before and after the push.
Afterward, A* resumes where it left off. This process repeats until A* beats the dungeon.

The repair process assures that all dungeons are playable, though only 10 of 30 GAN dungeons and 16 of 30 pure graph grammar dungeons needed any repair. Per dungeon, the average number of rooms repaired was less than one for the 30 GAN dungeons and the 30 graph grammar dungeons. 

\subsection{Rogue-like Game} \label{section:roguelike}

%\begin{figure}[t]
%\centering
%\includegraphics[width=\columnwidth]{game-example.png}
%\caption{Game Interface. On top is a map of the dungeon. Below is the main playing area where subjects control the @ symbol. Status information is on the right: hearts remaining, keys, bombs, and items (raft can go here). Below is a log which describes what happens during each turn, such as the character killing an enemy or picking up an item. Below the log is a legend describing what each character symbol represents in the game.}
%\label{fig:game_interface}
%\end{figure}

To interact with the dungeons, a Rogue-like game was created in Java %(Fig.~\ref{fig:game_interface}) 
using the AsciiPanel library by Trystans\footnote{\url{http://trystans.blogspot.com/}}. The Rogue-like game emulates the gameplay in \emph{Legend of Zelda}. However, the game is turn-based and only features one enemy type. %(seen as a red `e' character) 
Many fancy items in Zelda are absent, but there are still bombs, and every level has a raft. %, but it may not be required to complete generated dungeons.

All actions are turn-based, so combat is simple. The player moves first and then the enemies. %, so the player attacks first. 
If an enemy is adjacent to the player, including diagonal to it, it will attack. Each enemy attack has a 50\% chance of hitting and subtracting a heart from the player. The player can only attack enemies in orthogonally adjacent positions, by pressing the appropriate arrow key. When an enemy blocks the avatar's movement, an arrow press is an attack instead of a move. When enemies are not adjacent to the player, they move toward it, but only if the player is within line of sight of 4 tiles. Otherwise, they move randomly. Enemies also move over water tiles.

%% The items were already explained in the LOZ section, so description here is shortened
Upon death, enemies sometimes drop a heart or a bomb. If the player with no bombs enters an empty room, enemies sometimes spawn so the player will be able to pick up bombs. There is at least one bomb-able wall in each dungeon.

%%%%%%%%%%%%%%%%%%%%%%%%%%%%%%%%%%%%%%%%%%%%%%%%%%%%%%%%%%%%%%%%%%%%%%%%%%%%%%%%
\section{Human Subject Study} \label{section:hss}

The method of dungeon generation described thus far (\texttt{Graph+GAN}) is evaluated by having humans compare it to two other types of dungeon: a graph grammar dungeon that does not use a GAN (\texttt{Graph}), and Dungeon 4-1 from \emph{Legend of Zelda} (\texttt{Original}). Whenever a \texttt{Graph} dungeon places a room, it is chosen randomly from the set of all rooms in the VGLC training set. \texttt{Graph} and \texttt{Graph+GAN} dungeons seen by each participant were different. The \texttt{Original} dungeon played by every participant was Dungeon 4-1, because it is sufficiently interesting to represent a meaningful comparison. Some earlier dungeons are simplistic in comparison, and many later dungeons are so large that having users play them would be too time consuming. Dungeon 4-1 is also ideal because its raft item allows players to traverse obstacles in a new way, whereas many of the special items in other dungeons are weapons that introduce combat mechanics difficult to emulate in the Rogue-like engine.

The study had 30 participants (university students, faculty, and staff). Each participant played through a dungeon of each of the three types in a different order (5 per each of 6 possible orders). After each dungeon, the participant took a survey ranking the dungeon on a 1--5 scale in various categories. After the second dungeon, users indicated which of the two were better in various respects, and after the third dungeon all three were ranked relative to each other. Participants also provided open-ended text responses at each stage.

%%%% I don't think we actually used this information
%Additionally, they were asked about their video game experience in relation to games like Legend of Zelda at the very end.

Players start each dungeon with 0 bombs, 0 keys, and 4 hearts. It was possible to die, in which case the user would start the dungeon over, but the game would be easier. The starting/max number of hearts would increase, as would the chance of defeated enemies dropping a heart pickup. After dying, the starting hearts would increase to 6, then 8, then 20. Unexpectedly, one participant did not finish one of the dungeons even with this many tries, and thus repeated the dungeon starting over at 4 hearts. The heart drop rate for defeated enemies started at 30\%, and increased with each death to 60\%, then 90\% for the remaining deaths.

% Lives > 2: HDR: 30%, BDR: 40%; Lives = 2: HDR: 60%, BDR; 40%; Lives = 1: HDR: 90%, BDR: 10%; Lives = 0: HDR: 90%, BDR: 40%

Source code for running the user trials as well as video of the trials is accessible here: \url{southwestern.edu/~schrum2/zeldagan.html}.

%%%%%%%%%%%%%%%%%%%%%%%%%%%%%%%%%%%%%%%%%%%%%%%%%%%%%%%%%%%%%%%%%%%%%%%%%%%%%%%%
\section{Results} \label{section:results}

Statistical analysis of numerical ratings and relative rankings is provided, as are objective measures of the novelty of rooms in each dungeon type. Qualitative user responses provide additional insight into the quantitative data.

\subsection{Numeric Participant Ratings}

\begin{figure*}[t]
\centering
\includegraphics[width=\textwidth]{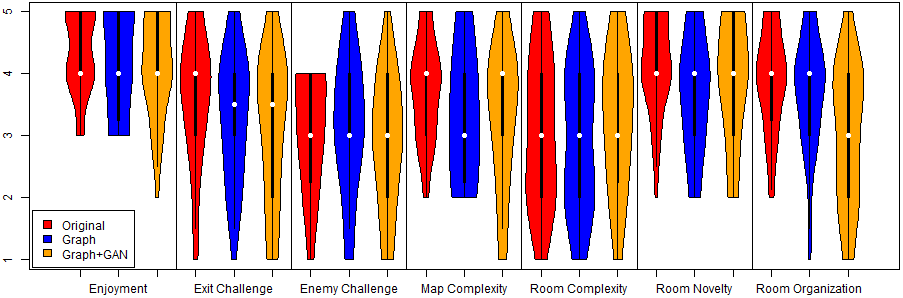}
\caption{Participant Ratings Of Each Dungeon Type. Violin plots depict distributions of participant ratings on a 1--5 scale for properties of each dungeon type. 
Each group of plots shows ratings of \texttt{Original}, \texttt{Graph}, and \texttt{Graph+GAN}.
The aspects being rated are under each group.
The only category with a statistically significant difference is \emph{Room Organization}: \texttt{Graph+GAN} rooms are less organized than others.}
\label{fig:userScores}
\end{figure*}

\texttt{Graph} and \texttt{Graph+GAN} dungeons are comparable to \texttt{Original} in most respects. Kruskal-Wallis tests ($df = 2$) indicate that there are no significant differences between dungeon types in terms of enjoyability ($H = 1.5065, p = 0.4708$), challenge in finding the exit ($H = 2.5478, p = 0.2797$), challenge from enemies ($H = 1.2331, p = 0.5398$), map complexity ($H = 2.8105, p = 0.2453$), room complexity ($H = 1.2279, p = 0.5412$), and room novelty ($H = 4.2023, p = 0.1223$). Only in terms of room organization is there a significant difference between dungeon types ($H = 11.337, p = 0.003454$), and post-hoc pairwise Mann-Whitney $U$ tests with FDR error correction show that it is specifically the \texttt{Graph+GAN} rooms that are less organized than rooms of both \texttt{Original} ($p = 0.0056$) and \texttt{Graph} ($p = 0.0164$). Since \texttt{Original} and \texttt{Graph} make use of the same set of rooms, there is no significant difference in their level of organization ($p = 0.4866$). Distributions of participant ratings for each dungeon type in all categories are shown as violin plots in Fig.~\ref{fig:userScores}.

%However, GAN and Graph grammar dungeons can produce a variety of levels as opposed to the 18 dungeons in the original game. Table \ref{tab:surveyResults} shows that while the original dungeon scored higher on average for each result, Graph Grammar and GAN dungeons are very close for almost every category. 

\subsection{Relative Participant Rankings of Dungeons}

\begin{figure*}[t]
\centering
\includegraphics[width=\textwidth]{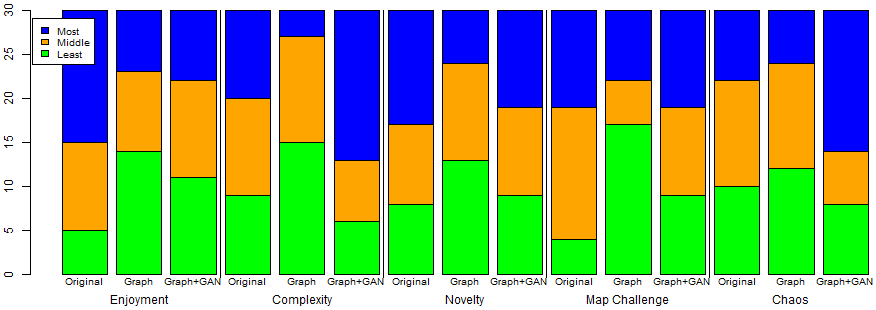}
\caption{Participant Relative Rankings Of Each Dungeon Type. Stacked bar charts show the number of participants that assigned each dungeon type a particular rank with respect to each other in each category. Categories are listed along the bottom. Each bar shows the count that ranked the given dungeon type as \emph{Least}, \emph{Middle}, and \emph{Most} from bottom to top. Some notable observations are that 15 participants rated \texttt{Original} as most enjoyable, 17 rated \texttt{Graph+GAN} as most complex, and 16 rated \texttt{Graph+GAN} as most chaotic. In contrast, 14 rated \texttt{Graph} as least enjoyable, 15 rated \texttt{Graph} as least complex, and 17 rated \texttt{Graph} map layouts as least challenging.}
\label{fig:userRanks}
\end{figure*}

After all three dungeons, participants ranked dungeons in terms of enjoyment, room complexity, room novelty, map layout challenge level, and chaos of the rooms (Fig.~\ref{fig:userRanks}). For each category the number 
of \emph{Most} and \emph{Least} ratings for each dungeon type were compared using exact multinomial tests. 

There is no significant difference in \emph{Most} ratings in the categories of enjoyment ($p = 0.185$), room novelty ($p = 0.2622$),  map challenge ($p = 0.7647$), or room chaos ($p = 0.07238$). The null hypothesis that was \emph{not} rejected is that the 30 user ratings are evenly split into 10 per dungeon type. Only for room complexity was there a significant difference between \emph{Most} ranks ($p = 0.005532$). Post-hoc pairwise binomial tests with FDR error correction indicate that \texttt{Graph+GAN} rooms received significantly more \emph{Most Complex} ranks than \texttt{Graph} (17 vs.\ 3, $p = 0.0077$),
though the ratings in Fig.~\ref{fig:userScores} indicate that the degree of the difference is small.
Also, despite \texttt{Graph} and \texttt{Original} rooms coming from the same set, there is no significant difference between the number of \emph{Most Complex} ranks of \texttt{Original} vs.\ \texttt{Graph+GAN} (10 vs.\ 17, $p = 0.2478$). The difference between \texttt{Original} and \texttt{Graph} was also not significant (10 vs.\ 3, $p = 0.1384$).

For \emph{Least} ranks, there was no significant difference in enjoyment ($p = 0.1117$), room complexity ($p = 0.156$), room novelty ($p = 0.5943$), or room chaos ($p = 0.0724$). However, there was a significant difference in map challenge ($p = 0.0184$). Specifically, post-hoc binomial tests with FDR correction indicate that \texttt{Graph} received significantly more \emph{Least Challenging} ranks than \texttt{Original} (17 vs.\ 4, $p = 0.022$). 
This finding is interesting because the layouts for \texttt{Graph+GAN} and \texttt{Graph} were defined by the same algorithm.
However, the differences between \texttt{Original} and \texttt{Graph+GAN} (4 vs.\ 9, $p = 0.267$) and \texttt{Graph} and \texttt{Graph+GAN} (17 vs.\ 9, $p = 0.253$) were not significant.

Despite few distinctions being statistically significant, there are interesting non-significant differences. First, 15 users found \texttt{Original} most enjoyable.
However, when compared with the 1--5 ratings in Fig.~\ref{fig:userScores}, it seems that the degree to which \texttt{Original} was more enjoyable was minor.
In contrast, 16 participants found \texttt{Graph+GAN} rooms most chaotic. The 1--5 ratings for \emph{Room Organization} relate to these responses, and indicate that GAN rooms may actually be moderately more chaotic/less organized. \texttt{Original} received the highest number of \emph{Most Novel} ranks (13) and smallest number of \emph{Least Novel} ranks (8) with respect to its rooms, whereas \texttt{Graph} received the most \emph{Least Novel} ranks (13) and least \emph{Most Novel} ranks (6). This contrast is strange because every room in \texttt{Original} is a room that could be in \texttt{Graph} dungeons. The GAN generated rooms, often unique to these dungeons, were ranked \emph{Most Novel} 11 times and \emph{Least Novel} 9 times. This confusion can be clarified with the objective measure of novelty presented next.

\subsection{Objective Novelty Comparisons}

An objective calculation of room novelty was made to measure differences between dungeon types. \emph{Room Novelty} is the average normalized distance of that room from all other rooms in its dungeon. The distance metric is the count of tile positions in which two rooms differ. Only the novelty of the primary $12 \times 7$ floor area is considered (excluding walls and doors). \emph{Dungeon Novelty} is the average novelty of all rooms in the dungeon. Summary novelty statistics are in Table \ref{tab:novelty}.

\setlength\tabcolsep{2pt}
\begin{table}[t]
\caption{\label{tab:novelty}Objective Novelty Scores.}

{\small Summary statistics of novelty scores for different collections of dungeons and rooms are shown. $N$ is the sample size. The first three rows are based on \emph{Dungeon Novelty}, and the next six on \emph{Room Novelty}. Calculations are performed across all rooms in the given collections, and across only the unique rooms. \texttt{Original} is less novel, unless you focus on unique rooms only.} 

\centering
\begin{tabular}{|r|c|c|c|c|}
\hline
Type & $N$ & Avg $\pm$ StDev & Min & Max \\
\hline % Images needed to be vertically centered within row
\texttt{Original} Dungeons & 18 & $0.2348 \pm 0.0496$ & 0.1311 & 0.3178 \\
\hline
\texttt{Graph} Dungeons & 30 & $0.2752 \pm 0.0485$ & 0.1975 & 0.3759 \\
\hline
\texttt{Graph+GAN} Dungeons & 30 & $0.2837 \pm 0.0357$ & 0.1970 & 0.3899 \\
\hline
\hline
All \texttt{Original} Rooms & 459 & $0.2481 \pm 0.1118$ & 0.1545 & 0.6733 \\
\hline
All \texttt{Graph} Rooms & 491 & $0.2941 \pm 0.0772$ & 0.2035 & 0.5920 \\
\hline
All \texttt{Graph+GAN} Rooms & 492 & $0.3019 \pm 0.0816$ & 0.2108 & 0.5891 \\
\hline
\hline
Unique \texttt{Original} Rooms & 38 & $0.3442 \pm 0.0888$ & 0.2453& 0.6062 \\
\hline
Unique \texttt{Graph} Rooms & 87 & $0.3437 \pm 0.0713$ & 0.2471 & 0.5334 \\
\hline
Unique \texttt{Graph+GAN} Rooms & 367 & $0.3268 \pm 0.0720$ & 0.2337 & 0.5802 \\
\hline
\end{tabular}
\end{table}

%The average novelty across all 18 dungeons from the original \emph{Legend of Zelda} is 0.2348. The least novel dungeon is 2-1 with a score of 0.1311, and the most novel dungeon is 9-2 with a score of 0.3178. 

%% ANOVA plus Tukey more appropriate
%Average novelty scores between different sets are compared using Welch's $t$-test because sample sizes differ. Here, \texttt{Original} refers to all 18 dungeons from the original game. Even though \texttt{Graph} only uses rooms from the original game, \texttt{Graph} is significantly more novel than \texttt{Original} ($t=2.7549, df=35.3, p = 0.009222$). \texttt{Graph+GAN} is also significantly more novel than \texttt{Original} ($t=3.6539, df=27.678, p = 0.001068$), but not significantly more novel than \texttt{Graph} ($t=0.7742, df = 53.328, p = 0.4422$). 

Comparing novelty scores of different dungeon types using one-way ANOVA reveals significant differences ($F(2,75)=7.317, p = 0.00125$). Post-hoc pairwise comparisons with Tukey's HSD and error-adjusted $p$-values are presented. Even though \texttt{Graph} only uses rooms from the original game, \texttt{Graph} is significantly more novel than \texttt{Original} ($p = 0.00854$). Here, \texttt{Original} refers to all 18 dungeons from the original game. \texttt{Graph+GAN} is also significantly more novel than \texttt{Original} ($p = 0.00116$), but not significantly different from \texttt{Graph} ($p = 0.7377$). 

The novelty of Dungeon 4-1 specifically is 0.2970, which is higher than the averages for all dungeon types; a very novel dungeon was used in this study. Users explicitly mentioned this: ``I enjoyed that the layout was different in almost all rooms.'' Fig.\ \ref{fig:tloz_4_1} verifies this, and indicates why users rated the novelty of this dungeon high, even though the set of all rooms in the original game has low novelty. 

In addition to calculating novelty scores for each dungeon, averages across all rooms present in a given collection of dungeons can also be calculated (Table \ref{tab:novelty}). ANOVA indicates a significant difference between the room novelties of all rooms from the original game, all rooms in all 30 \texttt{Graph} dungeons, and all rooms in all 30 \texttt{Graph+GAN} dungeons ($F(2,1439)=47.85, p < 0.0001$). Tukey's HSD once again indicates that \texttt{Graph} and \texttt{Graph+GAN} are significantly more novel than \texttt{Original} ($p < 0.0001$), but not significantly different from each other ($p = 0.3751$).

Calculations on sets of only the unique rooms of each collection are also performed (Table \ref{tab:novelty}), because these collections have many repeated rooms, especially those in \texttt{Original}. Although \texttt{Graph} uses the same rooms, they are sometimes modified by the repair process (Section \ref{section:repair}), so \texttt{Graph} has more unique rooms. \texttt{Graph+GAN} has the most unique rooms. When reduced to only unique rooms, there is no significant difference among types ($F(2,489)=2.517, p=0.0818$), indicating that \texttt{Original} dungeons re-use certain rooms more heavily than the random sampling of \texttt{Graph} or the GAN output of \texttt{Graph+GAN}.

\subsection{Informative Participant Quotes}

Quotes contextualize the quantitative findings. In particular, why was Dungeon 4-1 %from the original \emph{Legend of Zelda} 
appealing? Participants enjoyed the water obstacle that was only passable with the raft. One said, ``water cross tool/item was enjoyable.'' Another said, ``I liked that you had to wait later in the level to get the water walking thing and that helped you get further in the level.'' 

More generally, backtracking was appealing, as indicated about a \texttt{Graph} dungeon: ``I liked the need to backtrack through a couple of the dungeon rooms for necessary items if you didn't find them first.'' However, for the graph backbone in this study, only some generated levels required the raft to be beaten. In others, players found the raft before needing it. One user said of a \texttt{Graph} dungeon, ``I liked that this dungeon had rooms that used the raft more than the other; however, I got the raft early enough to where I didn't have to worry about water.'' 

Better design of the graph backbone could enforce backtracking as in Dungeon 4-1. However, some users appreciated how expectations were subverted: ``I liked that there was the raft item near the beginning of the dungeon that I could see but couldn't reach.  I felt like I had to figure out a way to get to the raft, but couldn't.'' Of a \texttt{Graph+GAN} dungeon: ``This dungeon was very chaotic, with items you didn't need in places you couldn't access. I liked that a lot because it threw me off and had me thinking about different possibilities.''

\begin{figure*}[t]
\centering
\includegraphics[width=0.135\textwidth]{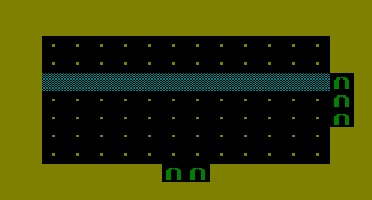}
\includegraphics[width=0.135\textwidth]{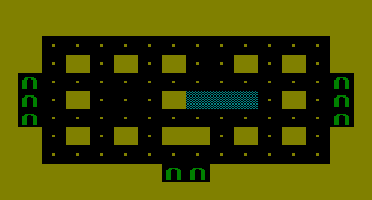}
\includegraphics[width=0.135\textwidth]{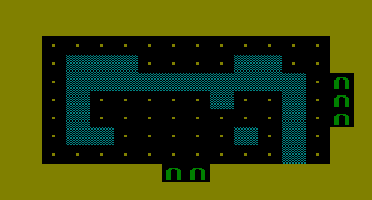}
\includegraphics[width=0.135\textwidth]{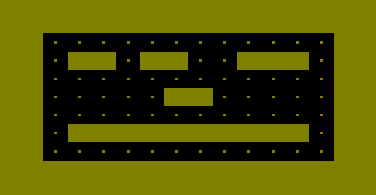}
\includegraphics[width=0.135\textwidth]{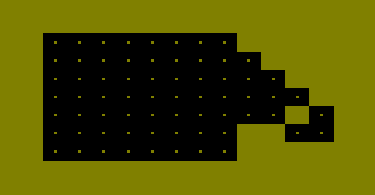}
\includegraphics[width=0.135\textwidth]{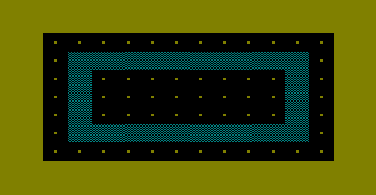}
\includegraphics[width=0.135\textwidth]{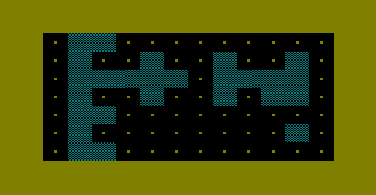}
\caption{Spectrum of Rooms Generated by the GAN. Some are identical or nearly identical to rooms in the training set, but others seem less structured and predictable, thus showcasing the diversity of the GAN outputs, but also revealing why its rooms are sometimes considered chaotic and unorganized.}
\label{fig:ganRooms}
\end{figure*}

This quote supports data indicating that GAN rooms are less organized. A participant observed: ``there were parts of rooms and enemies that I couldn't reach.'' This oddity could be avoided by restricting item and enemy placement to reachable locations. Reachability aside, many GAN rooms simply look more chaotic: ``There was a large mix of wall and water blocks, in ways that didn't seem completely natural. There was very little symmetry and a lot of obstacles.''

Although the GAN produces chaotic rooms, 10 participants specifically said things like ``They seemed organized,'' and ``I felt like the rooms were organized.'' The GAN also produces rooms from the original training set, and unique rooms that have a level of structure similar to original rooms (Fig.~\ref{fig:ganRooms}). Randomness led to some users seeing more rooms of one type than the other. Some people appreciated the chaotic rooms: ``they were chaotic but in a good way, none seemed like a copy of the previous and kept me on my toes.''

Much criticism was directed at \texttt{Graph} dungeon layouts. 
%Such quotes are not surprising given the significantly high number of \emph{Least Challenging} ranks assigned to \texttt{Graph}. 
Participants said, ``I didn't enjoy how simple the dungeon was overall,'' ``The map layout was very simple, not very novel,'' and ``this one favored simpler layouts.'' \texttt{Graph+GAN} dungeons did not receive many comments like this, despite using the same graph grammar. Randomness in generation may have played a role, though it may be that chaotic GAN room layouts distracted from issues with the dungeon layout. 

The most criticized layout was a linear layout without much branching: ``just a diagonal line, not many choices,'' and ``It was not as difficult to make it from room to room due to the lack of multiple bordering squares.'' These complaints could be remedied by having segments for the dungeon backbone with more diverse path options. However, the main issue seems to be randomness in the 2D layout, because some \texttt{Graph} dungeons \emph{had} interesting layouts: ``The map layout had me thinking of different areas the secret doors were in. It was interesting to try and figure out where to go next.''

Ultimately, conflicting opinions about several aspects of \texttt{Graph} and \texttt{Graph+GAN} dungeons are likely based partly on differing user preferences and perspectives, but are potentially also based on the variety of dungeons that can be produced by these methods, making it hard to categorize all dungeons of either type in the same way.

%%%%%%%%%%%%%%%%%%%%%%%%%%%%%%%%%%%%%%%%%%%%%%%%%%%%%%%%%%%%%%%%%%%%%%%%%%%%%%%%
\section{Discussion and Future Work} \label{section:discussionAndFW}

%TODO: Make several points: Zelda GAN produces enjoyable levels. However, regular graph grammar does to, and neither is actually significantly better than original level. Caveats: We only compared against one level that was pretty interesting ... in retrospect, retrieval of that ladder made things very interesting ... might not have been true with other levels in the game. Also, being as good as a level from such a classic, famous game is actually an achievement, but we don't just produce one level ... we can produce a nearly infinite selection of levels! But then why use Zelda GAN instead of plain graph grammar? Levels in the graph grammar are boring and repetitive. GAN had highest room complexity.

The \texttt{Graph+GAN} technique presented in this paper procedurally generates dungeons similar in terms of enjoyment, challenge level, and complexity to Dungeon 4-1 from \emph{The Legend of Zelda}. Dungeon 4-1 is special because it introduces the raft item which makes new types of puzzles possible. Creating dungeons comparable to this dungeon is impressive. Furthermore, the \texttt{Graph+GAN} technique can create an effectively infinite multitude of such dungeons.

%\todo[inline]{Some of the quotes can be interpreted as criticizing our design choices in how the graph grammar backbone was defined. For example, people thought that dead-end rooms had no purpose, but for us the purpose was to make the player get lost and waste effort, maybe even getting health chipped about by enemies. Some users also complained about there not being enough puzzle locked doors, or about the raft not being required to beat certain levels. In order to defend the strength of our level generation method, we have to insult our ability as level designers. There is no reason that a more well-designed graph grammar backbone could not accommodate the desires of these complaining users, but we did not provide the particular backbone they wanted ... therefore, a benefit of our method is that a small tweak to the backbone should provide the user with many levels they would like, whereas making a new hand-designed level to accommodate their wants and needs would be a major effort, or would result in a level too similar to the one they already played, but only fixing one specific issue with it.}

Improving the handcrafted backbone for the graph grammar could vastly improve layouts, remedying many user complaints. The dungeon generation method would be the same, but a better designer could encourage the method to produce better output. Tweaking the backbone requires relatively little effort, given that the benefit is an infinite multitude of levels. The backbone could be adjusted to force backtracking after obtaining the raft, and could provide any desired number of locked doors and/or puzzle rooms. Without a graph grammar, a designer can fix a specific level, but needs to expend great effort to create whole new levels adhering to a particular high-level design plan.

Both \texttt{Graph} and \texttt{Graph+GAN} techniques produce a multitude of levels, but \texttt{Graph} makes repetitive use of the same rooms. Even when it produces a layout as interesting as an \texttt{Original} level, it offers nothing new in terms of rooms. In contrast, GAN rooms are less organized, and considered most complex. Some users enjoyed the unpredictability of certain GAN rooms, but the GAN can also produce structured rooms similar to those from the original game.

%Between the Graph Grammar and GAN techniques, they are not much different. However, Graph Grammar is not very interesting as there are repetitive rooms and are not as complex as the GAN. The GAN rooms are significantly not as organized compared to the original rooms, making GAN rooms more unpredictable and interesting. However, some GAN rooms are not traversable without modification.

%Another thing to note about the generative methods, is the ability to modify the rooms in order to make the dungeons beatable, such as replacing walls with floor tiles. This can allow more variation between the rooms, which may explain the significantly higher novelty scores for both grammar methods compared to the original dungeons.

In the future, it is desirable to have a data-driven method replace the graph grammar entirely. Whether GANs or some other method can be adapted for this purpose is uncertain. The variation across the 38 unique rooms in the original game (training set) seems less than the variation across the 18 dungeons. There is a combinatorial explosion of potential complexity in complete dungeons when the variety of possible rooms is taken into account, and 18 dungeons of very different sizes may not be enough for a GAN to learn general design principles. However, bootstrapping methods (that work with limited data) for applying GANs to level design are an area of active research \cite{Torrado2019BootstrappingCG,park:cog19}. Generating the entire dungeon based on data will hopefully better capture design patterns of the original dungeons.

Rather than having a simple GAN generating walls and water blocks, it would be desirable to have generated enemies, keys, puzzle blocks, etc. For further research, a conditional GAN \cite{Mirza14conditionalGAN} could be used to generate rooms based on type. For instance, enemy rooms or puzzle rooms could be specifically requested. This approach would better match the original game's enemy, puzzle block, and key placement. 

Though this paper shows the potential of the \texttt{Graph+GAN} approach, a more impressive example would utilize all details of Zelda's levels, and create a gameplay experience closer to the original. Unfortunately, the VGLC data is lacking many details. However, the current GAN model could, without modification, generate rooms for a more intricate game if the gameplay engine were more complex. Tweaks to the engine could also enable online play, making the game more accessible for future studies.

%There is also a plan to work on a game that is more similar to the original Legend of Zelda for a better experience than the experimental one. No changes will be needed to the work mentioned in this paper as the generated dungeons can be used in numerous formats.

\section{Conclusions} \label{section:conclusion}

A new hybrid approach to generating game dungeons combining a Generative Adversarial Network with a Generative Graph Grammar was presented and validated with a user study. User responses indicate that results were comparable to a handcrafted level from \emph{The Legend of Zelda}. Better design of the graph backbone, and a more sophisticated game engine could result in a more impressive experience.
This new approach to Procedural Content Generation could prove valuable for commercial video games.

%However, our GAN generated dungeon proved to be novel as we are generating unique rooms each time. Additionally, the generated dungeon with rooms from the original were modified to ensure that the dungeon was beatable, which may result in an increase of novelty. Overall, this proves that this dungeon technique may prove valuable for commercial video games.

%%%%%%%%%%%%%%%%%%%%%%%%%%%%%%%%%%%%%%%%%%%%%%%%%%%%%%%%%%%%%%%%%%%%%%%%%%%%%%%%
%\section*{APPENDIX}


\section*{ACKNOWLEDGMENTS}

This research is supported in part by the Summer Collaborative Opportunities and Experiences (SCOPE) program,
funded by various donors to Southwestern University.

%%%%%%%%%%%%%%%%%%%%%%%%%%%%%%%%%%%%%%%%%%%%%%%%%%%%%%%%%%%%%%%%%%%%%%%%%%%%%%%%

\bibliographystyle{IEEEtran}
\bibliography{ZeldaGAN}

\end{document}